\documentclass{article}
\usepackage{spconf,amsmath,graphicx}
\usepackage{subfigure}
\usepackage{CJKutf8}
\usepackage{multirow}
\usepackage{verbatim}
\usepackage{url}
\usepackage{amsmath}
\usepackage{diagbox}
\usepackage{caption}
\usepackage{amsfonts,amssymb}
\usepackage{subfigure}
\usepackage{graphicx}
\usepackage{bm}
\usepackage{multirow}
\usepackage[ruled,lined,commentsnumbered]{algorithm2e}
\usepackage{booktabs}
\usepackage{bbm}
\usepackage[inkscapelatex=false]{svg}
\usepackage{pifont}
\usepackage{stfloats}
\usepackage{multicol}
\usepackage{float}
\usepackage{array}
\newcommand{\PreserveBackslash}[1]{\let\temp=\\#1\let\\=\temp}
\newcolumntype{C}[1]{>{\PreserveBackslash\centering}p{#1}}
\newcolumntype{R}[1]{>{\PreserveBackslash\raggedleft}p{#1}}
\newcolumntype{L}[1]{>{\PreserveBackslash\raggedright}p{#1}}

\usepackage{mathrsfs}
\usepackage{hyperref}
\usepackage{booktabs}

\title{Glancing Future for Simultaneous Machine Translation}
\name{Shoutao Guo \textsuperscript{\rm 1,2} \qquad Shaolei Zhang \textsuperscript{\rm 1,2} \qquad Yang Feng \textsuperscript{\rm 1,2}$^{\ast}$ \thanks{ \ \ $^{\ast}$Corresponding author: Yang Feng.} }
\address{\textsuperscript{\rm 1} Key Laboratory of Intelligent Information Processing, \\ Institute of Computing Technology, Chinese Academy of Sciences (ICT/CAS) \\
\textsuperscript{\rm 2} University of Chinese Academy of Sciences, Beijing, China
}
\begin{document}
%
\maketitle
\begin{abstract}
Simultaneous machine translation (SiMT) outputs translation while reading the source sentence. Unlike conventional sequence-to-sequence (seq2seq) training, existing SiMT methods adopt the prefix-to-prefix (prefix2prefix) training, where the model predicts target tokens based on partial source tokens. However, the prefix2prefix training diminishes the ability of the model to capture global information and introduces forced predictions due to the absence of essential source information. Consequently, it is crucial to bridge the gap between the prefix2prefix training and seq2seq training to enhance the translation capability of the SiMT model. In this paper, we propose a novel method that glances future in curriculum learning to achieve the transition from the seq2seq training to prefix2prefix training. Specifically, we gradually reduce the available source information from the whole sentence to the prefix corresponding to that latency. Our method is applicable to a wide range of SiMT methods and experiments demonstrate that our method outperforms strong baselines\footnote{Code is available at \href{https://github.com/ictnlp/Glance-SiMT}{https://github.com/ictnlp/Glance-SiMT}}.


\end{abstract}
\section{Introduction}
\label{sec:intro}
Simultaneous machine translation (SiMT) \cite{gu2017learning, ma2019stacl, ma2019monotonic}, which outputs the translation while reading the source sentence, is widely used in streaming scenarios, such as live broadcast and online conferences. Compared to conventional full-sentence translation \cite{vaswani2017attention}, it needs to predict the target tokens based on partial source tokens, thus facing greater challenges. 

The full-sentence translation employs the sequence-to-sequence (seq2seq) training for end-to-end translation \cite{sutskever2014sequence}. In contrast, the SiMT model adopts the prefix-to-prefix (prefix2prefix) training \cite{ma2019stacl} to accommodate the streaming inputs. However, this prefix2prefix training weakens the ability of the SiMT model to acquire global information from the source sentence \cite{zhang2022reducing}. More importantly, it may cause forced prediction, where the SiMT model predicts target tokens even when the essential source information is lacking during training \cite{chen2021improving}. This exacerbates the hallucinations of the SiMT model during inference. Therefore, bridging the gap between the prefix2prefix training and the seq2seq training becomes pivotal in augmenting the translation capability of the SiMT model and reducing hallucinations in translation.

\begin{figure}[t]
    \centering
    \includegraphics[width=3.0in]{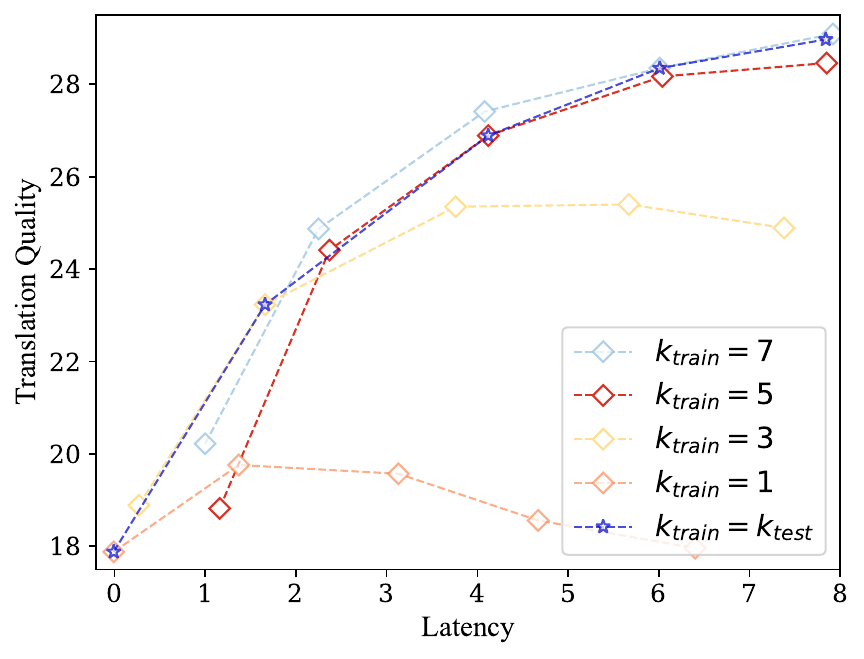}
    \caption{Translation performance of the SiMT model when performing different wait-$k$ policies ($k_{test}\in\{1,3,5,7,9\}$) on WMT15 De$\rightarrow$En dataset. A larger $k_{test}$ results in the greater latency. The wait-$k$ policy \cite{ma2019stacl} starts translation after reading $k$ tokens. The $k_{train}$ and $k_{test}$ represent the settings of $k$ for the wait-$k$ model during training and inference, respectively.    }
    \label{pre_alalysis}
\end{figure}

The existing SiMT methods commonly adopt the prefix2prefix training, and improve the performance of SiMT models by determining the appropriate source prefix for the target tokens \cite{Zhang2022WaitinfoPB}. According to the way to determine the prefix of the source sentence, the SiMT methods can be broadly categorized into fixed and adaptive policies. The fixed policy \cite{ma2019stacl, zhang2021future, zhang2021universal} generates translation using predefined heuristic rules and offers simplicity and stability as its distinguishing features. In contrast, the adaptive policy \cite{guo2022turning, zhang2022hidden, zhang2022information, zhang-feng-2023-end} allows the SiMT model to autonomously decide on the source prefix for each target token during translation. However, existing methods primarily focus on enabling the SiMT model to adjust to the prefix2prefix training rather than rectifying the inherent limitations of the prefix2prefix training.

A promising direction is to narrow the gap between the prefix2prefix training and the seq2seq training, in order to alleviate the deficiencies of the prefix2prefix training and enhance the capability of the SiMT model. In SiMT, the prefix2prefix training exhibits a closer resemblance to the seq2seq training under high latency compared to low latency scenarios \cite{zhang2022modeling}. To assess the feasibility of mitigating the limitations of the prefix2prefix training, we employ a SiMT model trained under high latency for translation tasks under low latency conditions. As depicted in Figure \ref{pre_alalysis}, when performing the wait-$1$ policy, the SiMT model trained with high latency (i.e., $k_{train}$=$3$) outperforms its low latency counterpart (i.e., $k_{train}$=$1$). However, SiMT models trained with even higher latency (i.e., $k_{train}$=$5$) experience a decrease in performance \cite{zhang2021universal}. This suggests that reducing the gap between the prefix2prefix and seq2seq training can enhance the capability of the SiMT model without deviating too far from the intended latency scenarios. Nonetheless, it is important to note that this approach is heuristic and offers limited improvements.

In this paper, we propose the \emph{glancing future training}, which glances future source information in curriculum learning \cite{bengio2009curriculum} to achieve the transition from the seq2seq training to prefix2prefix training. Specifically, we enable each target token to access the source tokens beyond its latency range. In the initial stage, the SiMT model is trained using the seq2seq training to enhance the translation capability. With the model being trained, we gradually reduce the future information available to the SiMT model to align it with the prefix2prefix training. Throughout the training process, each target token interacts with the source prefixes of different lengths, thereby enhancing the ability of the SiMT model to implicitly embed future information. By utilizing the glancing future training, our method enhances the ability of the SiMT model to utilize global information and alleviate the hallucinations. Our method is applicable to a wide range of SiMT methods and experiments on two SiMT translation tasks demonstrate that our method outperforms strong baselines in both fixed and adaptive policies.

\begin{figure*}[t]
\centering
\subfigure[Seq2Seq Training]{
\includegraphics[width=2.35in]{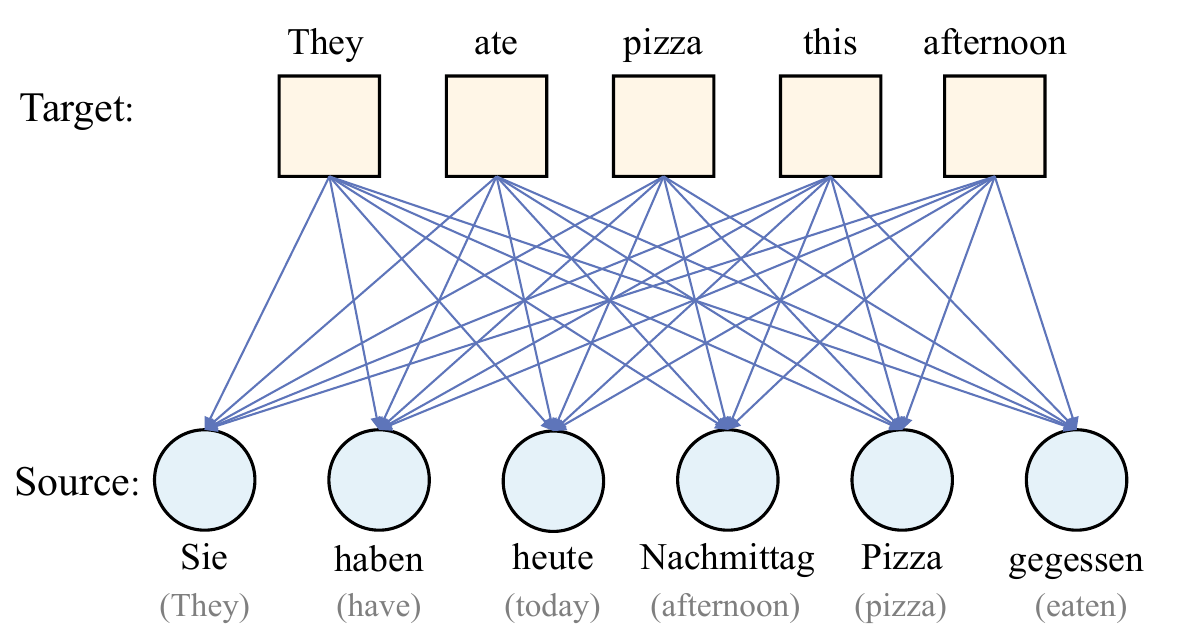}
\label{seq2seq}
}\hspace{-0.3cm}
\subfigure[Prefix2Prefix Training]{
\includegraphics[width=2.04in]{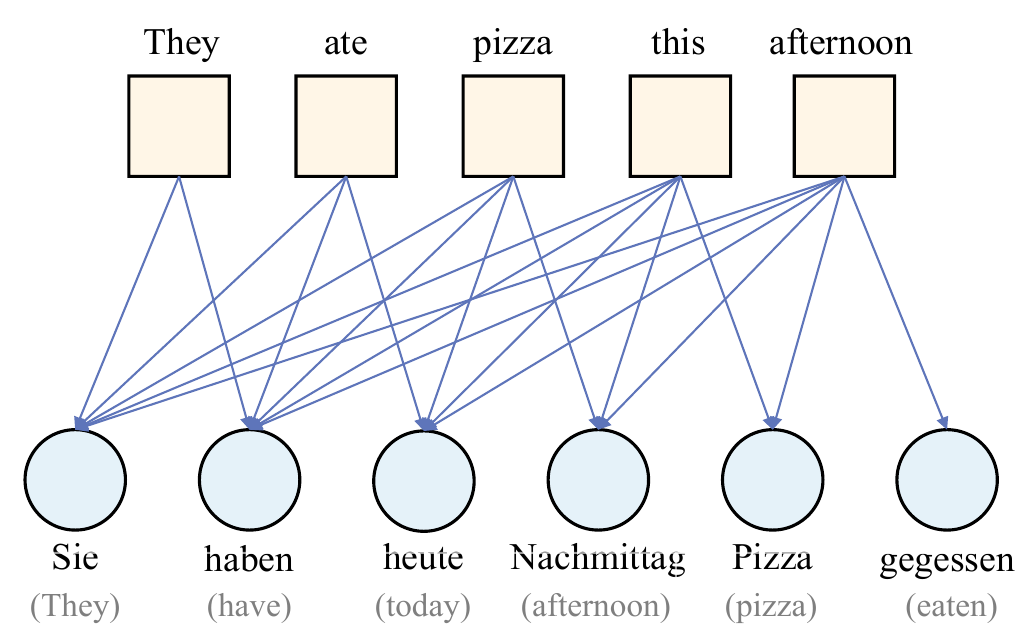}
\label{prefix2prefix}
}\hspace{-0.3cm}
\subfigure[Glancing Future Training]{
\includegraphics[width=2.04in]{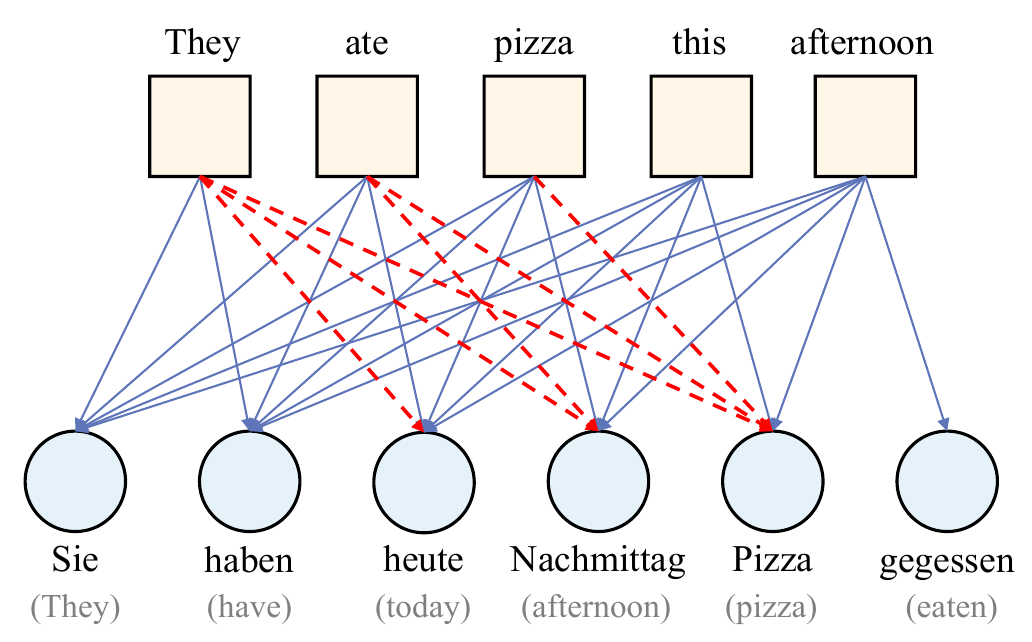}
\label{glancing}
}

\caption{A German$\rightarrow$English example under different training frameworks. The SiMT model is trained to learn wait-$2$ policy \cite{ma2019stacl} using the prefix2prefix training and glancing future training. $\alpha$ in our method is set to 0.75. The solid blue line represents the attention allowed by the policy. The red dashed line represents the extra attention allowed by our method. }
\label{main_fixed}
\end{figure*}

\section{Method}
\label{sec:method}
Our method attempts to glance future in curriculum learning to achieve the transition from the seq2seq training to prefix2prefix training. We will successively introduce the glancing future training and its integration with existing SiMT methods.

\subsection{Glancing Future Training}
Given the source sentence $\mathbf{x}$ = $(x_1, ... , x_J)$ with length $J$, the SiMT model generates the target sentence $\mathbf{y}$ = $(y_1, ..., y_I)$ with length $I$ according to the translation policy. In SiMT, the translation policy guides the model in determining the source prefix suitable for the translation of each target token \cite{guo-etal-2023-learning}. It can be formalized as $\mathbf{g}$ = $(g_1, ..., g_I)$, where $g_i$ represents the number of tokens in the source prefix when translating $y_i$. During training, the SiMT model is trained using the prefix2prefix training by minimizing the cross-entropy loss \cite{ma2019stacl}:
\begin{equation}
\mathcal{L}_{simt} = - \sum\limits_{i = 1}^{I} \log p(y^{\star}_i \mid \mathbf{x}_{\leq g_i}, \mathbf{y}_{<i}),
\end{equation}
where $y^{\star}_i$ denotes the ground-truth token. 

However, the prefix2prefix training diminishes the ability of the model to capture global information and may introduce forced predictions \cite{chen2021improving}. We introduce the \emph{glancing future training}, which can compensate for the problem of prefix2prefix training by exposing future source information to the SiMT model. To narrow the gap with seq2seq training, we make the target token access source tokens that exceed the limitation of latency during training. To this end, we define $f_i$ ($f_i \in [0, ..., J-g_i]$), which represents the number of source tokens that $y_i$ can pay extra attention to during training. Therefore, the optimized translation policy during training can be adjusted as follows:
\begin{equation}
    \hat{g}_i = g_i + f_i ,
\end{equation}
where $\hat{g}_i$ denotes the number of source tokens available to the SiMT model in our method during training.

Therefore, the key lies in determining $f_i$. As analyzed in Section \ref{sec:intro}, it is of help to mitigate the shortcomings of the prefix2prefix training by giving the SiMT model future information. However, too much future information can hinder the model from adapting to the latency scenarios, leading to a performance collapse \cite{zhang2022gaussian}. In response to this challenge, we introduce curriculum learning into the process of capturing future information. At the beginning of training, we let the model attend to the whole source sentence to enhance the translation capability. Then we gradually reduce the available source information to make the model adapt to the incomplete source inputs. To achieve this, we define $\alpha$, which controls the proportion of source information exposed to the SiMT model:
\begin{equation}
    \alpha = \alpha_{min} + (1-\alpha_{min}) \times \max(1 - \frac{N_{update}}{d}, 0),
\end{equation}
where $N_{update}$ is update steps, and $d$ is the hyperparameter to control the decaying degree. $\alpha_{min}$ is the minimum amount of future source information. Therefore, $f_i$ can be calculated as:
\begin{equation}
    f_i = (J-g_i) \times \alpha,
\end{equation}
where $J$ denotes the length of source sentence. Throughout the training, each target token gets the opportunity to interact with the prefixes of different lengths, enhancing the ability of the model to implicitly embed future source information \cite{qian2021glancing}.

\subsection{Integration into SiMT Method}
After introducing the general form of our method, we proceed to combine it with existing SiMT methods. Specifically, we select wait-$k$ and HMT policies \cite{zhang2022hidden} as the representatives of fixed and adaptive policies, respectively. We will briefly outline the integration of these policies with our method.

The wait-$k$ policy \cite{ma2019stacl} generates translation according to the predefined rules. During training, the SiMT model is trained to learn that policy under prefix2prefix training, as depicted in Figure \ref{prefix2prefix}. It initially reads in $k$ source tokens and then alternates between outputting and reading a token. Therefore, wait-$k$ policy can be formalized as:
\begin{equation}
    g^{wait\text{-}k}_i = \min \{k+i-1, J \},
\end{equation}
where $J$ denotes the length of the source sentence. Then the wait-$k$ policy can be trained using our glancing future training, as illustrated in Figure \ref{glancing}.

Hidden Markov Transformer (HMT) \cite{zhang2022hidden} is the state-of-the-art SiMT method. During training, it assigns $N$ hidden events $\mathbf{g}^{hmt}_i$ = $(g^{hmt}_{i,1}, ... , g^{hmt}_{i,N})$ to target token $y_i$. The event $g^{hmt}_{i,n}$ represents the number of source tokens in $n$-th event when translating $y_i$. Therefore, each target token can consider multiple hidden events and select the appropriate hidden event for translation. Therefore, $g^{hmt}_{i,n}$ can be calculated as:
\begin{equation}
    g^{hmt}_{i,n} = \min\{L + (i-1) + (n-1), J\},
\end{equation}
where $L$ denotes the source tokens read in before translating the target token $y_1$. Accordingly, $g^{hmt}_{i,n}$ is integrated into glancing future training in a manner similar to $g_i$.


\section{Experiments}
\subsection{Datasets}
We evaluate our method on IWSLT15\footnote{\href{https://nlp.stanford.edu/projects/nmt/}{https://nlp.stanford.edu/projects/nmt/}} English$\rightarrow$Vietnamese (En$\rightarrow$Vi) and WMT15\footnote{\href{www.statmt.org/wmt15/}{www.statmt.org/wmt15/}} German$\rightarrow$English (De$\rightarrow$En) datasets. For En$\rightarrow$Vi task, we use TED tst2012 as the development set and TED tst2013 as the test set. The tokens whose frequency is less than 5 are replaced with $\left \langle unk \right \rangle$. For De$\rightarrow$En task, we use newstest2013 as the development set and newstest2015 as the test set. The De$\rightarrow$En texts are encoded by BPE \cite{sennrich2016neural}.
\subsection{System Settings}
We compare our method with previous SiMT methods and introduce them briefly. 
\begin{list}{\labelitemi}{\leftmargin=1em} \vspace{-0.2cm}
\setlength{\itemsep}{0pt}
\setlength{\parsep}{0pt}
\setlength{\parskip}{0pt}
    \item {\bf Full-sentence} is the conventional full-sentence translation model based on Transformer \cite{vaswani2017attention}.

    \item {\bf Wait-$k$} policy reads $k$ tokens first and alternates between outputting and reading one token \cite{ma2019stacl}.

    \item {\bf MoE Wait-$k$}, which is the current state-of-the-art fixed policy, makes each head perform different Wait-$k$ policies and integrates the decisions of all heads \cite{zhang2021universal}.

    \item {\bf Adaptive-Wait-$k$} uses heuristic method to switch between different wait-$k$ models for translation \cite{Zheng2020SimultaneousTP}.

    \item {\bf MMA} makes each head determine the translation policy by predicting the Bernoulli variable \cite{ma2019monotonic}.

    \item {\bf LEAPT} allows the SiMT model to learn how to translate source sentence prefixes and utilize the future context \cite{lin2023leapt}.

    \item {\bf HMT} belongs to adaptive policy and achieves the state-of-the-art performance in SiMT \cite{zhang2022hidden}.

    \item {\bf Glance-Wait-$k$} applies our method on Wait-$k$.
    \item {\bf Glance-HMT} applies our method on HMT.
    
\end{list} \vspace{-0.2cm}
Our systems are adapted from Fairseq Library \cite{ott2019fairseq}. The hyperparameter $\alpha_{min}$ is set to 0.05. $d$ is set to 160000 on De$\rightarrow$En task and 8000 on En$\rightarrow$Vi task. Unless mentioned, other system settings are consistent with \cite{ma2019monotonic} and \cite{zhang2022hidden}. We use greedy search during inference and evaluate all methods with translation quality measured by BLEU \cite{papineni2002bleu} and latency estimated by Average Lagging (AL) \cite{ma2019stacl}.



\begin{figure*}[t]
\begin{minipage}[t]{.64\linewidth}
\vspace{-2mm}
\subfigure[En$\rightarrow $Vi]{
\includegraphics[width=2.2in]{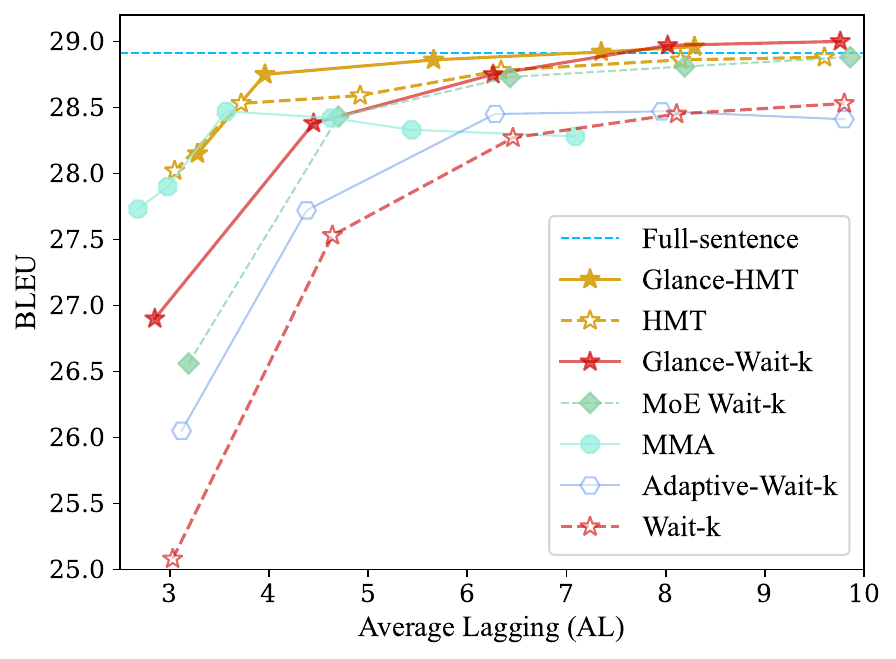}
}
\subfigure[De$\rightarrow $En]{
\includegraphics[width=2.1in]{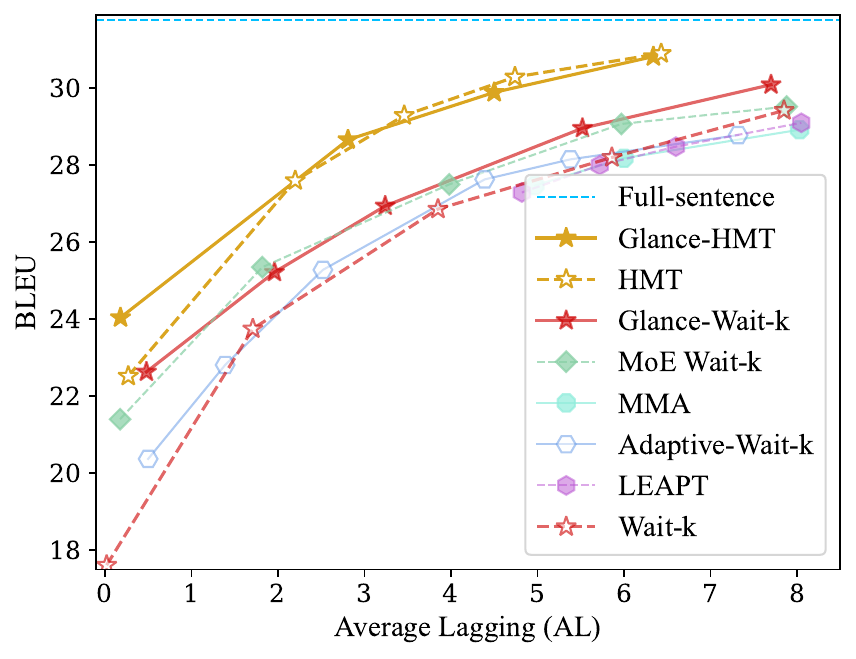}
}
\vspace{-2.5mm}
\caption{Translation performance of SiMT methods on En$\rightarrow$Vi and De$\rightarrow$En. }
\label{main_results}
\end{minipage}
\hspace{5mm}
\begin{minipage}[t]{.31\linewidth}
\vspace{0mm}
\centering
\includegraphics[width=2.14in]{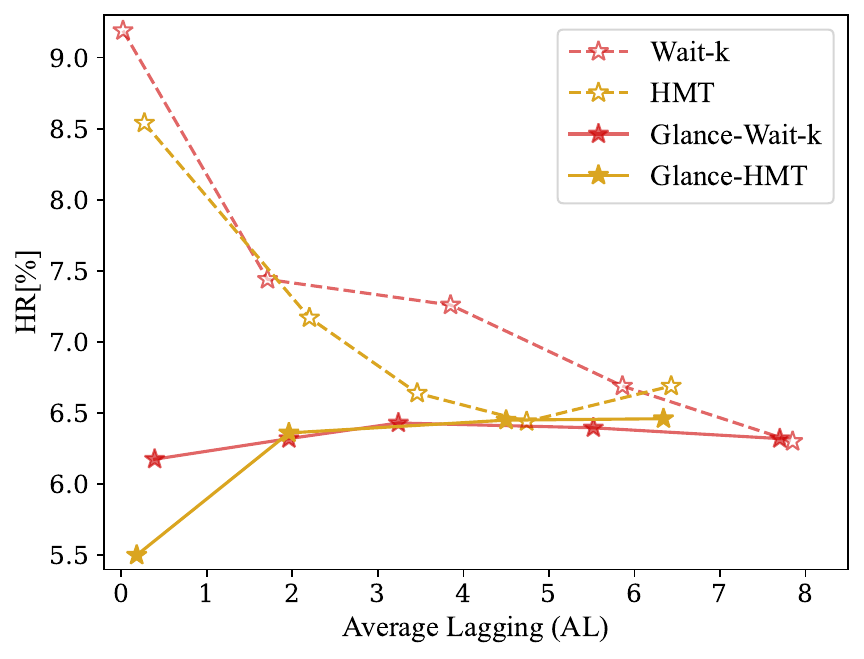}

\caption{Hallucination rate (HR) of different SiMT methods on De$\rightarrow$En task.}
\label{hr}
\end{minipage}

\end{figure*}

\begin{table}[]
\centering
\begin{tabular}{l c c  c} \toprule[1.2pt]
\textbf{$\;\;\;\;\;\;\;\;\;\;\;\;\;$Method} & $\alpha_{min}$ & \textbf{AL} & \textbf{BLEU}                                       \\ \cmidrule(lr){1-1} \cmidrule(lr){2-2}\cmidrule(lr){3-4}

 & 0.00 & -0.09 & 19.71 \\ 
Glance-Wait-$k$ & \textbf{0.05} & \textbf{0.39} &  \textbf{22.30} \\
 & 0.20 & 0.17 & 21.40 \\

\cmidrule(lr){1-1}\cmidrule(lr){2-2}\cmidrule(lr){3-4}

$\;\;\;\;$w/o Curriculum Learning & 0.05 & 0.4 & 21.74 \\

 \bottomrule[1pt]
\end{tabular}
\caption{Ablation study of Glance-Wait-$k$ on De$\rightarrow$En task. `w/o Curriculum Learning' denotes that the SiMT model is given $\alpha_{min}$ of future information throughout training process.}
\label{train_met}
\end{table}

\subsection{Main Results}
The performance comparison between our method and other methods is illustrated in Figure \ref{main_results}. Our method outperforms strong baselines in fixed and adaptive policies.

For fixed policy, Glance-Wait-$k$ demonstrates performance comparable to MoE Wait-$k$, which is recognized as the best fixed policy. Compared to Wait-$k$ policy, our method brings significant improvement, especially under low latency. While Wait-$k$ policy relies on predefined rules \cite{ma2019stacl}, our method empowers target tokens to interact with prefixes of varying lengths, thereby augmenting the ability to implicitly embed future information.

For adaptive policy, Glance-HMT achieves substantial improvements over HMT under low latency and delivers performance on par with HMT under high latency. As the best SiMT method, HMT possesses the capability to adapt its policy to enhance translation quality \cite{zhang2022hidden}. However, it is still influenced by the limitations of prefix2prefix training. Our method facilitates a transition from straightforward to difficult policies by gradually moving from full-sentence to low-latency scenarios. Therefore, our method empowers the adaptive method to devise better policies.


\subsection{Ablation Study}
To investigate the influence of different settings on our method, we conduct ablation studies on De$\rightarrow$En task. As illustrated in Table \ref{train_met}, curriculum learning brings significant improvements to our method, as it can empower each target token to interact with source prefixes of different lengths, thereby enhancing the translation ability of the SiMT model. The ablation study on $\alpha_{min}$ highlights the significance of supplying appropriate future information to the SiMT model \cite{zhang2021universal}, which is consistent with our analysis in Figure \ref{pre_alalysis}.

\begin{table}[]
\centering
\small
\begin{tabular}{c c c} \toprule[1.2pt]
\textbf{Method} & \textbf{AL} & \textbf{BLEU}                                       \\ \cmidrule(lr){1-1} \cmidrule(lr){2-3}

\textbf{Glance-Wait-$k$ (Adjacency)} & \textbf{0.39} & \textbf{22.30} \\ 
Glance-Wait-$k$ (Attention)  & 0.18 & 21.29 \\
Glance-Wait-$k$ (Randomization) & 0.22 & 21.56 \\

 \bottomrule[1pt]
\end{tabular}
\caption{Comparison of ways to capture future information on De$\rightarrow$En task. `Adjacency' signifies our method. `Attention' represents that the model selects future information with higher weights in cross-attention. `Randomization' denotes that the model randomly selects future information. }
\label{acquire_future}
\end{table}

\subsection{Analysis on Selecting Future Information}
We also analyze the selection of future information in our method. Table \ref{acquire_future} shows 
our method of selecting adjacent future information is more effective than other settings. We think that our method may provide more relevant future information to the given source prefix, thereby enhancing the translation capability of the SiMT model.

\subsection{Hallucination in Translation}
If the SiMT model is trained using the prefix2prefix training, it may be forced to predict target tokens without necessary source information during training. Consequently, the SiMT model is likely to generate hallucinations during inference \cite{Guo2023LearningOP}. To quantify the hallucinations in translation, we introduce the hallucination rate (HR) \cite{chen2021improving}. As depicted in Figure \ref{hr}, our method proves effective in reducing the hallucinations in translations. Through our glancing future training, we bridge the gap between the prefix2prefix training and the seq2seq training, thereby enhancing the translation capability.

\section{Conclusion}
In this paper, we propose a novel method, which glances the future in curriculum learning to bridge the gap between prefix2prefix training and seq2seq training. Experiments demonstrate that our method outperforms strong baselines.

\vfill\pagebreak
\bibliographystyle{IEEEbib}
\bibliography{strings,refs}

\end{document}